\documentclass{article}
\usepackage{spconf,amsmath,graphicx}
\usepackage{url}
\usepackage{balance}
\usepackage{array}
\newcolumntype{P}[1]{>{\centering\arraybackslash}p{#1}}
\newcolumntype{M}[1]{>{\centering\arraybackslash}m{#1}}
\newcommand{\at}{\makeatletter @\makeatother}

\title{A Simple but effective BERT Model for Dialog State Tracking on Resource-Limited Systems}
\name{Tuan Manh Lai$^{\dagger \star}$ \qquad Quan Hung Tran$^{\dagger}$ \qquad Trung Bui$^{\dagger}$ \qquad Daisuke Kihara$^{\star}$}

\address{$^{\star}$ Purdue University, West Lafayette, IN \\
		$^{\dagger}$Adobe Research, San Jose, CA}
\begin{document}
%
\maketitle
\begin{abstract}
In a task-oriented dialog system, the goal of dialog state tracking (DST) is to monitor the state of the conversation from the dialog history. Recently, many deep learning based methods have been proposed for the task. Despite their impressive performance, current neural architectures for DST are typically heavily-engineered and conceptually complex, making it difficult to implement, debug, and maintain them in a production setting. In this work, we propose a simple but effective DST model based on BERT. In addition to its simplicity, our approach also has a number of other advantages: (a) the number of parameters does not grow with the ontology size (b) the model can operate in situations where the domain ontology may change dynamically. Experimental results demonstrate that our BERT-based model outperforms previous methods by a large margin, achieving new state-of-the-art results on the standard WoZ 2.0 dataset \footnote{We did not use the DSTC-2 dataset because its clean version is no longer accessible (\url{http://mi.eng.cam.ac.uk/~nm480/dstc2-clean.zip}). In addition, it has the exact same ontology as the WoZ 2.0 dataset.}. Finally, to make the model small and fast enough for resource-restricted systems, we apply the knowledge distillation method to compress our model. The final compressed model achieves comparable results with the original model while being 8x smaller and 7x faster.
\end{abstract}
\section{Introduction}
\label{sec:intro}
Task-oriented dialog systems have attracted more and more attention in recent years, because they allow for natural interactions with users to help them achieve simple tasks such as flight booking or restaurant reservation. Dialog state tracking (DST) is an important component of task-oriented dialog systems \cite{Young2013POMDPBasedSS}. Its purpose is to keep track of the state of the conversation from past user inputs and system outputs. Based on this estimated dialog state, the dialog system then plans the next action and responds to the user. In a \textit{slot-based} dialog system, a state in DST is often expressed as a set of slot-value pairs. The set of slots and their possible values are typically domain-specific, defined in a domain ontology.

With the renaissance of deep learning, many neural network based approaches have been proposed for the task of DST \cite{Mrksic2016NeuralBT,Liu2017AnET,zhong-etal-2018-global,Ren2018TowardsUD,nouri2018toward,mrkvsic2018fully,Korpusik2019DialogueST,chao2019bert}. These methods achieve highly competitive performance on standard DST datasets such as DSTC-2 \cite{Henderson2014TheSD} or WoZ 2.0 \cite{Wen2016ANE}. However, most of these methods still have some limitations. First, many approaches require training a separate model for each slot type in the domain ontology \cite{Mrksic2016NeuralBT,zhong-etal-2018-global,mrkvsic2018fully}. Therefore, the number of parameters is proportional to the number of slot types, making the scalability of these approaches a significant issue. Second, some methods only operate on a fixed domain ontology \cite{Liu2017AnET,zhong-etal-2018-global}. The slot types and possible values need to be defined in advance and must not change dynamically. Finally, state-of-the-art neural architectures for DST are typically heavily-engineered and conceptually complex \cite{zhong-etal-2018-global,Ren2018TowardsUD,nouri2018toward,mrkvsic2018fully}. Each of these models consists of a number of different kinds of sub-components. In general, complicated deep learning models are difficult to implement, debug, and maintain in a production setting.

Recently, several pretrained language models, such as ELMo \cite{Peters:2018} and BERT \cite{Devlin2019BERTPO}, were used to achieve state-of-the-art results on many NLP tasks. In this paper, we show that by finetuning a pretrained BERT model, we can build a conceptually simple but effective model for DST. Given a dialog context and a candidate slot-value pair, the model outputs a score indicating the relevance of the candidate. Because the model shares parameters across all slot types, the number of parameters does not grow with the ontology size. Furthermore, because each candidate slot-value pair is simply treated as a sequence of words, the model can be directly applied to new types of slot-value pairs not seen during training. We do not need to retrain the model every time the domain ontology changes. Empirical results show that our proposed model outperforms prior work on the standard WoZ 2.0 dataset. However, a drawback of the model is that it is too large for resource-limited systems such as mobile devices.

To make the model less computationally demanding, we propose a compression strategy based on the knowledge distillation framework \cite{Hinton2015DistillingTK}. Our final compressed model achieves results comparable to that of the full model, but it is around 8 times smaller and performs inference about 7 times faster on a single CPU.


\begin{figure}[!ht]
\centering
\includegraphics[width=0.46\textwidth]{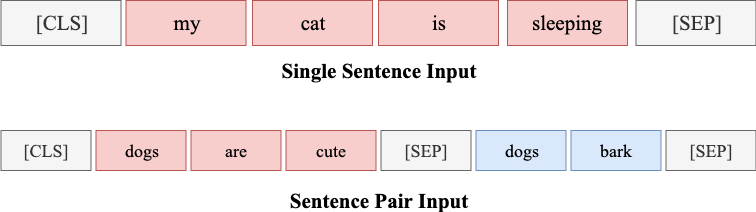}
\caption{BERT input format.}
\label{fig:bert_input_format}
\end{figure}

\section{Method}
\label{sec:method}
\subsection{BERT}
BERT is a powerful language representation model pretrained on vast amounts of unlabeled text corpora \cite{Devlin2019BERTPO}. It consists of multiple layers of Transformer \cite{Vaswani2017AttentionIA}, a self-attention based architecture. The base version of BERT consists of 12 Transformer layers, each with a hidden size of 768 units and 12 self-attention heads. The input to BERT is a sequence of tokens (words or pieces of words). The output is a sequence of vectors, one for each input token. The input representation of BERT is flexible enough that it can unambiguously represent both a single text sentence and a pair of text sentences in one token sequence. The first token of every input sequence is always a special classification token - $[$\texttt{CLS}$]$. The output vector corresponding to this token is typically used as the aggregate representation of the original input. For single text sentence tasks (e.g., sentiment classification), this $[$\texttt{CLS}$]$ token is followed by the actual tokens of the input text and a special separator token - $[$\texttt{SEP}$]$. For sentence pair tasks (e.g., entailment classification), the tokens of the two input texts are separated by another $[$\texttt{SEP}$]$ token. This input sequence also ends with the $[$\texttt{SEP}$]$ token. Figure \ref{fig:bert_input_format} demonstrates the input representation of BERT.

During pretraining, BERT was trained using two self-supervised tasks: masked language modeling (masked LM) and next sentence prediction (NSP). In masked LM, some of the tokens in the input sequence are randomly selected and replaced with a special token $[$\texttt{MASK}$]$, and then the objective is to predict the original vocabulary ids of the masked tokens. In NSP, BERT needs to predict whether two input segments follow each other in the original text. Positive examples are created by taking consecutive sentences from the text corpus, whereas negative examples are created by picking segments from different documents. After the pretraining stage, BERT can be applied to various downstream tasks such as question answering and
language inference, without substantial task-specific architecture modifications.

\begin{figure*}
\centering
\includegraphics[width=0.8\textwidth]{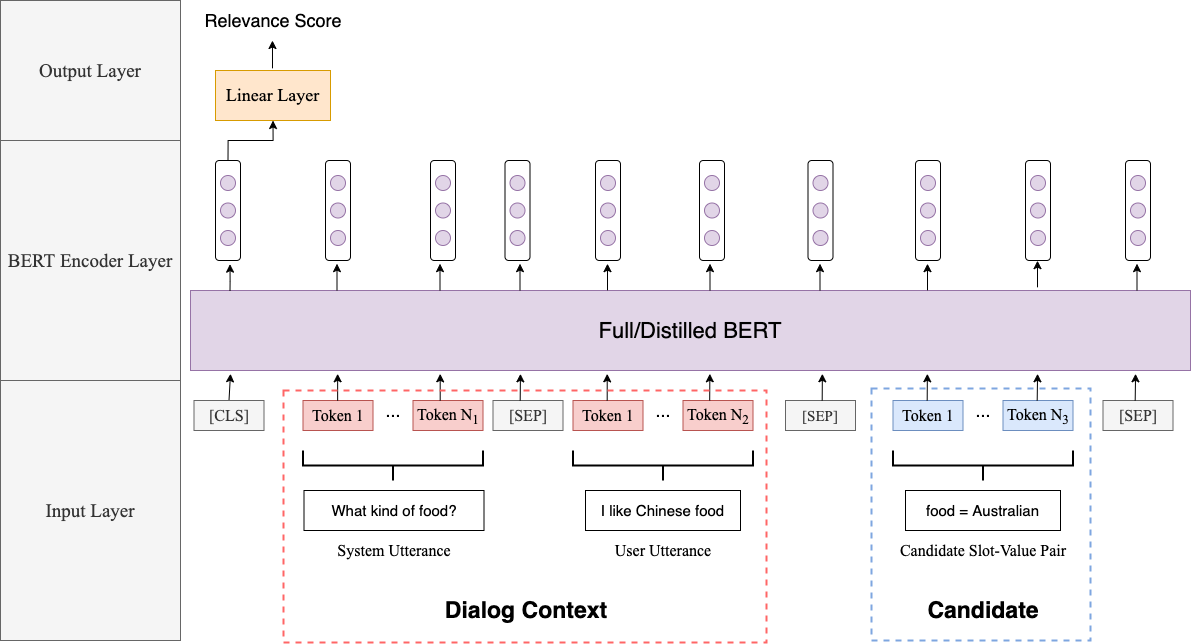}
\caption{Architecture of our BERT-based model for DST.}
\label{fig:general_architecture}
\end{figure*}

\subsection{BERT for Dialog State Tracking}
Figure \ref{fig:general_architecture} shows our proposed application of BERT to DST. At a high level, given a dialog context and a candidate slot-value pair, our model outputs a score indicating the relevance of the candidate. In other words, the approach is similar to a sentence pair classification task. The first input corresponds to the dialog context, and it consists of the system utterance from the previous turn and the user utterance from the current turn. The two utterances are separated by a $[$\texttt{SEP}$]$ token. The second input is the candidate slot-value pair. We simply represent the candidate pair as a sequence of tokens (words or pieces of words). The two input segments are concatenated into one single token sequence and then simply passed to BERT to get the output vectors $(\textbf{h}_1, \textbf{h}_2 \cdots ,\textbf{h}_M)$. Here, $M$ denotes the total number of input tokens (including special tokens such as $[$\texttt{CLS}$]$ and $[$\texttt{SEP}$]$).

Based on the output vector corresponding the first special token - $[$\texttt{CLS}$]$ (i.e., $\textbf{h}_1$) the probability of the candidate slot-value pair being relevant is:
\begin{equation}
y = \sigma(\textbf{W}\textbf{h}_1 + \textbf{b}) \in {\rm I\!R}
\end{equation}
where the transformation matrix $\textbf{W}$ and the bias term $\textbf{b}$ are model parameters, and $\sigma$ denotes the sigmoid function. It squashes the score to a probability between 0 and 1.

At each turn, the proposed BERT-based model is used to estimate the probability score of every candidate slot-value pair. After that, only pairs with predicted probability equal to at least 0.5 are chosen as the final prediction for the turn. To obtain the dialog state at the current turn, we use the newly predicted slot-value pairs to update the corresponding values in the state of previous turn. For example, suppose the user specifies a \texttt{food}$=$\texttt{chinese} restaurant during the current turn.  If the dialog state has no existing \texttt{food} specification, then we can add \texttt{food}$=$\texttt{chinese} to the dialog state. If \texttt{food}$=$\texttt{korean} had
been specified before, we replace it with \texttt{food}$=$\texttt{chinese}.

Compared to previous works \cite{zhong-etal-2018-global,Ren2018TowardsUD,nouri2018toward,mrkvsic2018fully}, our model is conceptually simpler. For example, in the GLAD model \cite{zhong-etal-2018-global}, there are two scoring modules: the utterance scorer and the action scorer. Intuitively, the utterance scorer determines whether the current user utterance mentions the candidate slot-value pair or not, while the action scorer determines the degree to which the slot-value pair was expressed by the previous system actions. In our proposed approach, we simply use a single BERT model for examining the information from all sources at the same time.

\subsection{Model Compression}
BERT is a powerful language representation model, because it was pretrained on large text corpora (Wikipedia and BookCorpus). However, the original pretrained BERT models are  computationally expensive and have a huge number of parameters. For example, the base version of BERT consists of about 110M parameters. Therefore, if we directly integrate an existing BERT model into our DST model, it will be difficult to deploy the final model in resource-limited systems such as mobile devices. In this part, we describe our approach to compress BERT into a smaller model.

Over the years, many model compression methods have been proposed \cite{Ba2013DoDN,Han2015DeepCC,Han2015LearningBW,Hinton2015DistillingTK}. In this work, we propose a strategy for compressing BERT based on the knowledge distillation
framework \cite{Hinton2015DistillingTK}. Knowledge distillation (KD) aims at transferring knowledge acquired in one model (a teacher) to another model (a student) that is typically smaller. We generally assume that the teacher has previously been trained, and that we are estimating parameters for the student. KD suggests training by matching the student's predictions to the teacher's predictions. In other words, we train the student to mimic output activations of individual data examples represented by the teacher.

We choose the pretrained base version of BERT as the teacher model. Our student model has the same general architecture as BERT but it is much smaller than the teacher model (Table \ref{table:model_size}). In the student model, the number of Transformer layers is 8, each with a hidden size of 256 units and 8 self-attention heads. The feedforward/filter size is 1024. Overall, our student model has 14M parameters in total and it is 8x smaller and 7x faster on inference than our original teacher model.

We first extract sentences from the BooksCorpus \cite{Zhu2015AligningBA}, a large-scale corpus consisting of
about 11,000 books and nearly 1 billion words. For each
sentence,  we use the WordPiece tokenizer \cite{Schuster2012JapaneseAK,Devlin2019BERTPO} to tokenize the sentence into a sequence of tokens. Similar to the pretraining phase of BERT, we mask 15\% of the tokens in the sequence at random. After that, we define the cross-entropy loss for each token as follow:
\begin{equation}
    \mathcal{L}(\textbf{a}_\text{T}, \textbf{a}_\text{S}) = \mathcal{H} \Bigg(\text{softmax} \bigg ( \frac{\text{\textbf{a}}_\text{T}}{\tau} \bigg ), \text{softmax} \bigg ( \frac{\text{\textbf{a}}_\text{S}}{\tau} \bigg ) \Bigg)
\end{equation}
where $\mathcal{H}$ refers to the cross-entropy function, $\textbf{a}_\text{T}$ is the teacher model's pre-softmax logit for the current token, $\textbf{a}_\text{S}$ is the student model's pre-softmax logit. Finally, $\tau$ is the temperature hyperparameter \cite{Hinton2015DistillingTK}. In this work, we set $\tau$ to be 10. Intuitively, $\mathcal{L}(\textbf{a}_\text{T}, \textbf{a}_\text{S})$ will be small if the student's prediction for the current token is similar to that of the teacher model. The distillation loss for the entire sentence is simply defined as the sum of all the cross-entropy losses of all the tokens. To summarize, during the KD process, we use this distillation loss to train our student model from scratch using the teacher model’s logits on unlabeled examples extracted from the BooksCorpus. After that, we can integrate our distilled student BERT model into our DST model (Figure \ref{fig:general_architecture}) and use the final model for monitoring the state of the conversation.

Different from the very first work on exploring knowledge distillation for BERT \cite{Tang2019DistillingTK}, our approach does not use any data augmentation heuristic. We only extract unlabeled sentences from the BooksCorpus to build training examples for distillation. Our work is in spirit similar to DistilBERT \cite{Sanh2019DistilBERTAD}, which also uses the original BERT as the
teacher and a large-scale unlabeled text corpus as the basic learning material. However, as shown in Table \ref{table:model_size}, the DistilBERT model is about 5 times larger than our student model. Recently, at WWDC 2019, Apple presented a BERT-based on-device model for question answering \cite{AppleCoreML}. Instead of using knowledge distillation, Apple used the mixed precision training technique \cite{Micikevicius2017MixedPT} to build their model. From the Table \ref{table:model_size}, we see that the model of Apple is much larger than our student model, as it has 8x more parameters and requires 4x more storage space. This implies that our student model is small enough to be deployed on mobile systems. To the best of our knowledge, we are the first to explore the use of knowledge distillation to compress neural networks for DST.

\section{Experiments and Results}

\begin{table}[ht!]
\small
\centering
 \begin{tabular}{|p{0.26\textwidth} | P{0.08\textwidth} | P{0.06\textwidth} |} 
 \hline
 Model & Number parameters & Storage Size\\ [0.5ex] 
 \hline
 Teacher Model ($N=12$, $d_1=768$, $d_2=3072$, $h=12$)&  110M & 440MB\\
 \hline
 Student Model ($N=8$, $d_1=256$, $d_2=1024$, $h=8$) & \textbf{14M}  & \textbf{55MB} \\
 \hline\hline
 Apple Core ML BERT \cite{AppleCoreML} & 110M &  220 MB\\
 \hline
 DistilBERT \cite{Sanh2019DistilBERTAD} ($N=6$, $d_1=768$, $d_2=3072$, $h=12$) & 66M & 270 MB\\
 \hline
 GLAD \cite{zhong-etal-2018-global} &  17M & --- \\
 \hline
\end{tabular}
\caption{Approximated size of different models. For models based on BERT, $N$ is the number of Transformer layers, $d_1$ is the hidden size, $d_2$ is the feedforward size, and $h$ is the number of self-attention heads}
\label{table:model_size}
\end{table}
\label{sec:exps_and_results}
\subsection{Data and Evaluation Metrics}
To evaluate the effectiveness of our proposed approach, we use the standard WoZ 2.0 dataset. The dataset consists of user conversations with dialog systems designed to help users find suitable restaurants. The ontology contains three \textit{informable} slots: \texttt{food}, \texttt{price}, and \texttt{area}. In a typical conversation, a user would first search for restaurants by specifying values for some of these slots. As the dialog progresses, the dialog system may ask the user for more information about these slots, and the user answers these questions. The user’s goal may also change during the dialog. For example, the user may want an `expensive' restaurant initially but change to wanting a `moderately priced' restaurant by the end. Once the system suggests a restaurant that matches the user criteria, the user can also ask about the values of up to eight \textit{requestable} slots (\texttt{phone}, \texttt{address}, ...). The dataset has 600/200/400 dialogs for train/dev/test split. Similar to previous work, we focus on two key evaluation metrics introduced in \cite{Henderson2014TheSD}: \textit{joint goal accuracy} and \textit{turn request accuracy}.

\subsection{Dialog State Tracking Results}
\begin{table}[ht!]
\small
\centering
 \begin{tabular}{|c | c | c|} 
 \hline
 Model & Joint goal & Turn Request \\ [0.5ex] 
  \hline
  \textbf{Full BERT-based model} & \textbf{90.5}  & 97.6\\
  \hline
  \textbf{Distilled BERT-based model} & 90.4  & \textbf{97.7}\\
  \hline
  \hline
 StateNet \cite{Ren2018TowardsUD} & 88.9 & --- \\
 \hline
 GCE \cite{nouri2018toward} & 88.5 & 97.4 \\
 \hline
 GLAD \cite{zhong-etal-2018-global} & 88.1 &  97.1 \\
 \hline
  BERT-DST PS \cite{chao2019bert} & 87.7 & --- \\
 \hline
  Neural Belief Tracker - CNN \cite{Mrksic2016NeuralBT} & 84.2 & 91.6  \\
  \hline
  Neural Belief Tracker - DNN \cite{Mrksic2016NeuralBT} & 84.4 & 91.2 \\ 
  \hline
\end{tabular}
\caption{Test accuracies on the WoZ 2.0 restaurant reservation datasets.}
\label{table:final_results}
\end{table}
Table \ref{table:final_results} shows the test accuracies of different models on the WoZ 2.0 dataset. Our full BERT-based model uses the base version of BERT (i.e., the teacher model in the knowledge distillation process), whereas the distilled BERT-based model uses the compressed student model. Both our full model and the compressed model outperform previous methods by a considerable margin. Even though our compressed model is 8x smaller and 7x faster than the full model (Table \ref{table:model_size}), it still achieves almost the same results as the full model. In fact, the smaller model has a slightly higher turn request accuracy. From Table \ref{table:model_size}, we see that our compressed model even has less parameters than GLAD, a DST model that is not based on BERT. This demonstrates the effectiveness of our proposed knowledge distillation approach.

Note that BERT-DST PS \cite{chao2019bert} is a recent work that also utilizes BERT for DST. However, the work focuses only on situation where the target slot value (if any) should be found as word segment in the dialog context. According to Table \ref{table:final_results}, our models outperform BERT-DST PS on the WoZ dataset. Furthermore, BERT-DST PS only uses the original version of BERT, making it large and slow.

\begin{table}[ht!]
\small
\centering
 \begin{tabular}{|c | P{0.08\textwidth} | P{0.08\textwidth} |} 
 \hline
 Model & Inf. Time on CPU (secs) & Inf. Time on GPU (secs)\\ [0.5ex] 
 \hline
 Full BERT-based Model &  1.465 & 0.113\\
 \hline
 \textbf{Distilled BERT-based Model} &  \textbf{0.205} & \textbf{0.024} \\
  \hline
 DistilBERT \cite{Sanh2019DistilBERTAD} & 0.579 & 0.043 \\
 \hline
\end{tabular}
\caption{Inference time of different models on CPU and GPU. We measured the average time it takes for each model to process one dialog turn (in seconds).}
\label{table:inference_speed_cpu}
\end{table}

\subsection{Size and inference speed}
Table \ref{table:model_size} shows that our distilled student model is much smaller than many other BERT-based models in previous works. Table \ref{table:inference_speed_cpu} shows the inference speed of our models on CPU (Intel Core i7-8700K \at 3.70GHz) and on GPU (GeForce GTX 1080). On average, on CPU, our compressed model is 7x faster on inference than our full model and 3x faster than DistilBERT. On GPU, our compressed model is about 5x faster than the full model and 2x faster than DistilBERT.

\section{Conclusion}
\label{sec:conclusion}
In this paper, we propose a simple but effective model based on BERT for the task of dialog state tracking. Because the original version of BERT is large, we apply the knowledge distillation method to compress our model. Our compressed model achieves state-of-the-art performance on the WoZ 2.0 dataset while being 8x smaller and 7x faster on inference than the original model. In future work, we will experiment on more large scale datasets such as the MultiWOZ dataset \cite{Budzianowski2018MultiWOZA}. 

\balance

\bibliographystyle{IEEEbib}
\bibliography{refs}

\end{document}